\documentclass[journal]{journal}
\ifCLASSINFOpdf
\else
\fi
\usepackage[]{algorithm2e}
\usepackage{algorithmic}
\usepackage{graphicx}
\begin{document}
%
\title{Trolls Identification within an Uncertain
Framework}
%
%
%

\author{Imen~Ouled~Dlala,~\IEEEmembership{}
        Dorra~Attiaoui,~\IEEEmembership{}
				Arnaud~Martin,~\IEEEmembership{}
        and~Boutheina~Ben~Yaghlane~\IEEEmembership{}
\thanks{Manuscript received.}}

%
%

\markboth{Journal of \LaTeX\ Class Files,~Vol.~6, No.~1, January~2007}%
{Shell \MakeLowercase{\textit{et al.}}: Bare Demo of IEEEtran.cls for Journals}
%



\maketitle
\thispagestyle{empty}

\begin{abstract}
The web plays an important role in people's social lives since the emergence
of Web 2.0. It facilitates the interaction between users, gives them the possibility to freely
interact, share and collaborate through social networks, online communities forums, blogs,
wikis and other online collaborative media. However, an other side of the web is negatively taken such as posting inflammatory messages. Thus, when dealing with the online communities forums, the managers seek to always enhance the performance of such platforms. In fact, to keep the serenity and prohibit
the disturbance of the normal atmosphere, managers always try to novice users against these malicious persons by
posting such message (DO NOT FEED TROLLS). But, this kind of warning is not enough
to reduce this phenomenon. In this context we propose a new approach for detecting
malicious people also called 'Trolls' in order to allow community managers to take their ability to post
online. To be more realistic, our proposal is defined within an uncertain framework. Based on the assumption consisting on the trolls' integration in the successful discussion
threads, we try to detect the presence of such malicious users.
Indeed, this method is based on a conflict measure of the belief function theory applied between the different messages of the thread. In order to show the feasibility and the result of our approach, we test it in different simulated data.
\end{abstract}


%
\IEEEpeerreviewmaketitle

\section{Introduction}
\IEEEPARstart{T}{he} way we look for, and acquire information has shifted greatly into to
instant, easy and low cost process. In fact, thanks to the Internet one can make a research
in any given topic, get a huge amount of information by a simple click. 
 Although, for some problems it is difficult to get satisfactory answers by searching
directly on a traditional search engine. Instead, we prefer to find someone who has expertise or
experience.
In order to have the best answer, one of the tools that has widened the scope of information exchange is Question Answering Communities (Q\&AC). 
These systems allow everyone to contribute as much as they can on a given community. Unfortunately, not all messages can be considered as reliable: some users claim themselves as experts, and other people post messages without any utility for the one who is seeking for answers. 
 Thus, the managers of these communities seek to always enhance the performance of such platforms. 
Although, the increase of the useless messages can be attributed to the presence of trolls. The term of trolling has been defined in several works within different communities, including  [2], [6] and [17]. These Malicious people intend to insidiously mislead the subject of the discussion in order to provoke controversy and disrupt the discussion. They aim to make normal users fall into their traps by deviating them from the main topic of the discussion. In fact, the only way to deal with a troll is to ignore him, or detect his presence in order to notify him or take away his ability to post online. Thus, some other works tried to detect not just their characteristics but also their presence in order to avoid them. To address this problem Cambria et al. [14] proposed a technique based on semantic computing to automatically detect and check web trolls. This work aims to prevent the malicious people from emotionally hurting other users or communities within the same social network. In another work Ortega et al. [15] proposed a method to classify users in a social network regarding to their trustworthiness. The goal of their method is to detect trolls from the other users by preventing such malicious users to gain high reputation in the
network. Patxi et al. [16] dealt with Trolling users on twitter social network. These studies were explored in different social networks within certain framework. 

When dealing with real-world applications, the massive amounts of data are inseparably
connected with imperfection. In fact, this kind of data can be imprecise
and/or uncertain or even missing.
Different theories have emerged to deal with this kind of data such as fuzzy set theory [21], possibility theory [22] and belief function theory [1]. Thus, to be closer to reality and to obtain more relevant results, we propose a new method dealing with uncertain data. This method aims to detect trolls in Q\&AC using the framework of belief function. 

The paper proceeds as follows: in Section 2, we introduce the Q\&AC and briefly review related works. In section 3, we present the necessary background regarding the different concepts of the belief functions theory. We define the different steps of our proposal based on a conflict measure in section 4. Finally, we present the feasibility of the proposed method on an illustrated example.

\section{Q\&AC: Quick overview}
In this section we introduce some concepts related the Q\&AC.
First we will start by presenting the main actors in these forums, then a little overview on sources identification and finally the levels of uncertainty we can face in Q\&AC.
\subsection{Users within Q\&AC}
Users are considered as the main actors within Question Answering Communities. We can define different types such as experts, trolls and learners.

\begin{itemize}
\item \textbf{Reliable user / Expert}: a person who is very knowledgeable about or skilful in a particular area.

\item \textbf{Troll}: a person who seeks to disturb the serenity of the concerned community. His purpose is to create controversial debates by multiplying irrelevant messages that we keep unanswered. 

\item \textbf{Learner}: a normal user of the Question Answering Community, trying to gain information and expertise.

\end{itemize}

\subsection{Sources Identification within Q\&AC}
Several researches have been exploring this field, trying to evaluate sources
of information in Q\&AC. Such as Bouguessa et al. [7] who proposed a model to identify authoritative
users based on the number of best answers provided by them. A best answer is
selected either by the asker or by other users via a voting procedure.
In [12], the author focused on the selection of questions a user would choose for
answering. Based on these studies, experts prefer answering questions where they
have a higher chance of making a valuable contribution.
Recently in [13], the authors proposed a framework for evaluating both the reliability and the expertise of an information provider. Considering some cognitive and
behavioral criteria of the users, they were able to establish a trust system. Using
a response matrix summarizing the interactions between peers of persons, each one
is capable of estimating and providing an opinion. Using the subjective logic to aggregate
these evaluations, they provided later a global reliability and expertise value for each
user within Q\&AC.
\subsection{Uncertainty within Q\&AC}
When dealing with information provided by humans, we are facing several levels of uncertainty. Gjergji et al. proposed three levels for Q\&AC [11], the first one is related to the extraction and integration of uncertainty, the second  deals with information sources uncertainty and finally the inherent knowledge related to the information itself.
In our case, we are more interested in the evaluation of the sources and the part of uncertainty related to them. The main issue in these communities is that we are facing users that we do not always have an \textit{apriori} knowledge about them. We ignore every thing about the sources' credibility, reliability, relevance, objectivity and expertise. In this context, we will exploit all the mathematical background and large panel of sepcificities provided by the theory of belief functions to help us considering this problem in an uncertain point of view.

\section{Theory of Belief Functions}
\label{sec:greetings}
This section recalls the necessary background related to the belief function theory 
It has been developed by Dempster in his work on upper and lower probabilities
[1]. Based on that, he was able to represent more precisely the observed data.

A belief function must take into consideration all the possible events on which a
source can describe a belief. Based on that, we can define the frame of discernment.

\subsection{Frame of discernment}
It is a finite set of disjoint elements noted $\Omega$ where $\Omega= \{\omega_{1},...,\omega_{n}\}$. This theory allows us to affect a mass on a set of hypotheses not only a singleton like in the probabilistic theory. Thus, we are able to represent ignorance, imprecision...

\subsection{Basic belief assignment $(bba)$}

A $bba$ is defined on the set of all subsets of $\Omega$, named power set and noted $2^{\Omega}$.
It affects a real value from $ [0, 1]$ to every subset of $ 2^{\Omega}$ reflecting sources amount of belief on this subset.
A $bba$ $m$ verifies:

\begin{eqnarray}
\sum_{X\subseteq \Omega} m(X) = 1.
\end{eqnarray}

We consider any positive elementary mass $m(X)>0$ as a focal element such that X belongs to $2^\Omega$.
\subsection{Combination rules}
Many combination rules have been proposed taking in consideration the nature of the sources.

\subsubsection{Dempster's combination rule}
The first one was proposed by Dempster in 1967 [1] which is a conjunctive normalized combination rule also called the orthogonal sum. 
Given two mass functions $m_{1}$ and $m_{2}$, for all $X \in 2^{\Omega}$, $X\neq \emptyset$, the Dempster's rule is defined by:
\begin{eqnarray}
 m_{D}(X)=m_1\cap m_2(X)= \frac{1}{1-k} \sum_{Y_{1} \cap Y_{2}=X}  m_{1}(Y_{1})m_{2}(Y_{2})
\end{eqnarray}

where $k=\sum_{Y_{1}\cap Y_{2}=\emptyset}  m_{1}(Y_{1})m_{2}(Y_{2})$ is the inconsistency of the fusion (or of the combination) can also be called the conflict or global conflict. $(1-k)$ is the normalization factor of the combination in a closed world.

\subsubsection{The conjunctive combination rule}

In order to consider the issues of the open world, the conjunctive combination rule was introduced by Smets [9].
Considering two mass functions $m_{1}$ and $m_{2}$, for all $X \in 2^{\Omega}$ $m_{conj}$ is defined by:

\begin{eqnarray}
 m_{conj}(X)= \sum_{Y_{1}\cap Y_{2}=X}  m_{1}(Y_{1})m_{2}(Y_{2})
\end{eqnarray}



\subsubsection{The disjunctive combination rule}

First introduced by Dubois and Prade 1986 [18], the induced results of two bbas  $m_{1}$ and $m_{2}$ is defined as follows:
\begin{eqnarray}
\forall X \subseteq \Omega m_{disj}(X)= \sum_{Y_{1}\cup Y_{2}=X} m_{1}(Y_{1}) m_{2}(Y_{2})
\end{eqnarray}

The disjunctive combination rule can be used when one of the sources is reliable or when we have no knowledge about their reliability.

\section{Inclusion as a conflict measure for belief functions}

Recently Martin in [3] used a degree of inclusion as involved in the measurement made in order to determine the conflict during the combination of two belief functions. He presented an index of inclusion having binary values where:
\begin {equation}
 \label{}
Inc(X_{1},Y_{2}) = \left\{
    \begin{array}{ll}
        \ 1, \mbox{if} X_{1} \subseteq Y_{2} \\
        \ 0, \mbox{otherwise}
    \end{array}
\right.
\end {equation}

With $X_{1}$, $Y_{2}$ being respectively the focal elements of $m_{1}$ and $m_{2}$.
This index is then used to measure the degree of inclusion of the two mass functions and defined as:

\begin{eqnarray}
   d_{inc}= \frac{1}{|F_{1}| |F_{2}|} \sum_{X_{1}\in F_{1}} \sum_{Y_{2}\in F_{2}}  Inc(X_{1},Y_{2})
  \end{eqnarray}
	Where $|F_{1}|$ and $|F_{2}|$ are the number of focal elements of $m_1$ and $m_2$. He define the degree of inclusion of $m_1$ and $m_2$: $\sigma_{inc}(m_{1},m_{2})$ as follows:
   \begin{eqnarray}
   \sigma_{inc}(m_{1},m_{2})= max(d_{inc}(m_{1},m_{2}),d_{inc}(m_{2},m_{1}))
  \end{eqnarray}

Where $d_{inc}$ is the degree of inclusion  of $m_{1}$ in $m_{2}$ and inversely. This inclusion is used as a conflict measure for two mass functions, using it like presented:
\begin{eqnarray}
Conf(m_{1},m_{2})=(1-\sigma_{inc}(m_{1},m_{2})d(m_{1},m_{2}))
  \end{eqnarray}
where $d(m_{1},m_{2})$, is the distance of Jousselme [10]:
\begin{eqnarray}
  d(m_{1},m_{2})=\sqrt{\frac{1}{2}(m_{1}-m_{2})^{T} \underline{\underline{D}} (m_{1}-m_{2})}
  \end{eqnarray}
 where $\underline{\underline{D}}$ is a metric based on the measure of Jaccard:
 \begin {equation}
 \label{}
D(A,B) = \left\{
    \begin{array}{ll}
        \ 1,if A=B=\emptyset \\
        \frac{|A \cap B|}{|A \cup B|}, \forall A,B \in 2^\Omega
    \end{array}
\right. \\ 
\end {equation}
\section{Trolls Identification based in a conflict measure}

Based on the assumption that consists of the trolls' integration in the successful discussion threads, we propose a new method for detecting malicious people in online communities forums. This approach is defined within the framework of belief functions. Indeed, it is based on a conflict measure of this theory applied between the different messages of the thread. We can summarize our proposed method in three major steps that will be discussed in depth in the following.

\subsection{Users' messages}
Hardarker proposed primary characteristics of a troll [2] (Aggression, Deception, Disruption, Success).
In 2014, Buckels et al. [6] specified the behavioral characteristics of a troll. They described them as persons having sadism, psychopathy and machiavilism.
To them, trolling is a "deceptive, destructive or disruptive manner in social media". 
\\
In the context of this work, to distinguish between the troll and the other users, we tried to manually
extracted the characteristics of their responses from the answers and comments in different forums. Based  on these characteristics, the content of the messages can be:
Off-topic, senseless or controversy. Using these characteristics, we have defined the frame of discernment that can characterize a message in a forum: 
\begin{eqnarray}
\Omega_{msg}= \left\{Off-topic, Senseless,1,\ldots,N \right\}
\end{eqnarray}

\begin{itemize}
\item Senseless: how much the response is empty of meaning?
\item Off-topic: How irrelative the answer can be?
\item $[1..N]$ : number of topics where, $[1..N] \backslash i$ with $i$ being the relevant topic, and $[1..N] \backslash i$ are the controversy topics posted by a troll.
\end{itemize}

During this step, we assume that a method of analysis expresses a piece of evidence concerning the nature of each message. This method aims to analyze the messages relative to the posted question or topic.

\subsection{Users' conflict}
Detecting irrelevant messages does not only means that this user is a troll. Thus, it is not only
the content of the messages that can characterize the trolls. We can find a victim user that
responds to a message posted by a troll. Besides, the subject of the discussion can change gradually.
In fact, to distinguish between trolls and other users in a community, we need to quantify
how a given user is in conflict with the rest of all the other users. Thus, we will base our approach on
measuring the conflict between the messages of each person posting answers. The list of notations is shown in table \ref{not}. 
\begin{table}[htbp]
\caption{List of notations}
\centering
\begin{tabular} {|c|p{5.5cm}|}
\hline
\textbf{Notations} & \textbf{Description} \\ \hline
$\textbf{U}$ & Users \\ \hline
$\textbf{N}$ & Number of users \\ \hline
$\textbf{NP}$ & Number of all the previous messages \\ \hline
$\bf{NP_j}$ & Number of the previous messages of a user $U_{j}$ \\   \hline
$\bf{N_{i}}$ & Number of all messages posted by a user $U_{i}$ \\   \hline
$\bf{N_{j}}$ & Number of all messages posted by a user $U_{j}$ \\   \hline
$\bf{m_{k}}$ & $k^{th}$ message of a user $U_{i}$   \\   \hline
$\bf{m_{s}}$ & $s^{th}$ message of a user $U_{j}$ \\    \hline
$\bf{Rank(m)}$ & Rank of the message m \\    \hline
$\bf{Tab1}$ & Contains in each time the conflict of a message relative to each user \\    \hline
$\bf{Tab2}$ & Contains in each time the number of the previous messages of a message \\    \hline
$\bf{Tab3}$ & Contains the total conflict of each user \\    \hline
$\bf{Conf_t}$ & Contains the sum of conflict of each user \\    \hline
\end{tabular}
\label{not}
\end{table}
 Using the inclusion as a conflict measure for belief functions, for each user $U_i$ we will measure:
\begin{itemize}
\item \textbf{$Conf_{msg_{/U}}$}: measures the conflict between the $k^{th}$ message posted by $U_{i}$ and the messages that were posted before it by each other users $U_{j}$.
\end{itemize}

\begin{center}
$ Conf_{msg_{/U}}(m_{k}(U_{i}),m(U_{j}))=$
\end{center}
\begin{equation}
\!\!\! \frac{1}{NP_j}\sum_{s=1}^{NP_j}  Conf(m_{k}(U_{i}),m_{s}(U_{j})), 
(i\neq j) \nonumber
\end{equation}
\begin{itemize}
\item \textbf{$Conf_{msg}$}: measures the conflict between the $k^{th}$ message posted by $U_{i}$ and the all messages that were posted before it by all the other users $U$ based on a weighted mean.
This measure takes into account the number of messages posted by every user in order to determine the level of conflict especially between a troll and an expert.
\end{itemize}

 \begin{center}
 $Conf_{msg}(m_{k}(U_{i}),m(U))= $
 \end{center}
\begin{equation}
\sum_{j=1}^{N} \frac{NP_j}{NP}  Conf_{msg_{/U}}(m_{k}(U_{i}),m(U_{j}))
\end{equation}
\begin{itemize}
\item \textbf{$Conf_{user}$}: measures the global conflict of the user $U_{i}$ 
\end{itemize}
\begin{equation}
 Conf_{user}= \frac{1}{N_i} \sum_{k=1}^{N_i}  Conf_{msg}(m_{k}(U_{i}),m(U)) \\
\label{eq}
\end{equation}


%
%
%
%
%
The value of the total conflict of a user can be risen when this user launches into an interminable debate with a troll. In this case, this victim user becomes in his turn a troll. Thus, the managers have to control the behavior of the users in many discussion threads.

\subsection{Users' clustering}
The last step consists on the classification of the users according to their conflict results into two groups. Therefore, to make decision we base our approach on an unsupervised classification method using the k means algorithm.

It was introduced by McQueen [19] and implemented in its current forms by Forgy [20]. The Kmeans algorithm aims to construct from the objects of the training set K partitions (clusters) concentrated and isolated from each other. In our case, we will devise the users into two partitions: K= 2. Since the value of the troll 's conflict is bigger than the conflict of the other users:

- Trolls belong to the group having the biggest value of center.

- The other users belong to the group having the least value of center.

\section{Experimentation}
To illustrate the comportment of our proposed method, we have tested it in different simulated data. In this section, we will present two different examples.  

\subsection{Example 1}
As we presented our method, it has three main steps. Indeed, we will present the results
of each step:
\subsubsection{Users' messages}
\begin{figure}[!t]

\includegraphics[height=5.5in,width=3.6in]{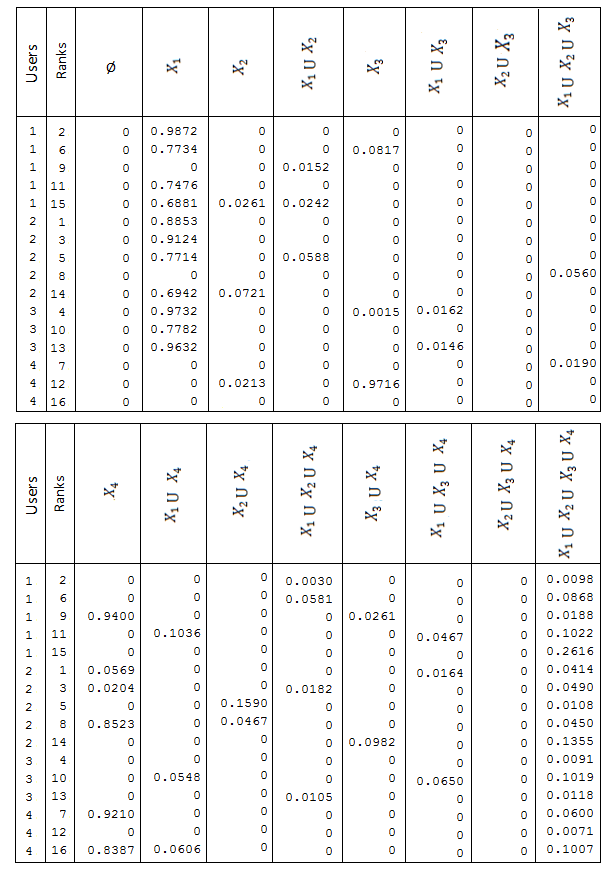}
\caption{Simulation Results}
\label{sim}
\end{figure}

Our assumption consists on the integration of the trolls in the successful discussion threads. From this point of view, we simulate the data of analyze of messages as depicted in Figure \ref{sim}. In fact, in this example we
will try to detect a troll in a group of 4 users. In this scenario, the discussion thread
contains 16 messages posted by different users and among whom three messages are
published by a troll.
In this example, our frame of discernment is composed by 4 elements: Relevant=$X_1$, off-topic=$X_2$, senseless=$X_3$, controversy-topic=$X_4$. As shown in Figure 2 each row presents: the owner of the message, the order of the message in the discussion thread and the mass function of this message (as mention in section \ref{sec:greetings}. B each bba must be equal to 1).

In this example, the first message of the troll ($U_4$) is controversy: $m(X_4)=0.9210$.
His second message is empty of meaning: $m(X_3)=0.9716$.
His third message is controversy: $m(X_4)=0.8387$.

\subsubsection{Users’ conflict}
Based on the method of the inclusion and applying our algorithm, we will present
the total conflict of each user of our example in Table \ref{table2}: $U_4$ has the biggest value of conflict. The total conflicts of users $U_1$ and U$_2$ is small relative to the total conflict of $U_4$ despite the fact that they responded to the first message of the troll by posting each one a controversy message. this result can be explained by the answers provided by these two users who have published relevant messages. 
$U_3$ has a small value of conflict, he published three relevant messages where in his first message $m(X_1)=0.9732$, in his second message  $m(X_1)=0.7782$, and his third message $m(X_1)=0.9632$.
\begin{table}[htbp]
\caption{Total conflict of each user}
\centering
\begin{tabular} {|l|l|l|l|l|}
\hline
 & $\bf{U_1}$&$\bf{U_2}$&$\bf{U_3}$&$\bf{U_4}$ \\ \hline
$\bf{Conf_{user}}$ & 0.0610&0.0639&$0.0489$&0.2030 \\ \hline

\end{tabular}
\label{table2}
\end{table}


\subsubsection{Users’ clustering}
Applying the $K$-means algorithm to the different values of total conflict of all users we obtained two clusters.

- Trolls$=\{ U_4\}$

- Other users$=\{ U_1, U_2, U_3\}$

Our proposal provides us a correct classification of the users. This result shows the feasibility of our proposed method.
\subsection{Example 2}
\begin{figure}[!h]
	\centering
		\includegraphics [width=6.5cm]{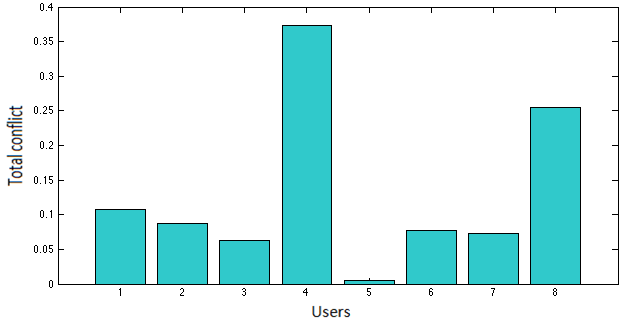}
		\caption{The steps of detecting trolls}
	\label{fig_sim2}
\end{figure}

For this simulation we will assume that we are dealing with 8 users, among them two trolls. The discussion thread contains 31 messages. The result of the total conflict of each user  expressed in equation \ref{eq} is illustrated in figure \ref{fig_sim2}. 

The first troll $U_4$ published 2 controversy messages and the second troll $U_8$ published 3 messages: The two first ones are off-topic, and  the last one is controversy. \\
- $U_1$ posted 3 relevant messages and 2 controversy messages to respond to the first troll.\\
- $U_2$ posted 7 relevant messages and 2 controversy messages to respond  to the first troll.\\
- $U_3$ posted 4 relevant messages and one off-topic message to respond to the second troll.\\
- $U_5$ posted one relevant message.\\
- $U_6$ published 3 relevant messages.\\
- $U_7$ published 2 relevant messages.

The total conflict of the troll $U_4$ is bigger relatively to the other troll $U_8$ because he published his posts after a big number of reliable messages provided by the other users. So, this situation created a higher value of a conflict. Applying the Kmeans algorithm our method provides us a correct classification:  

- Trolls$=\{ U_4, U_8\}$

- Other users$=\{ U_1, U_2, U_3, U_5, U_6, U_7\}$ 

 The users $U_1$, $U_2$ and $U_3$ are not classified among the trolls in spite of their posts that can be categorized as trolls' messages. This result is explained by the fact that they have other relevant messages. 

\section{Conclusion}

We proposed in this paper a new method for detecting 'Trolls' in Q\&AC. Relying on this approach managers, can control the behavior of the users in many discussion threads in order to notify them to stop trolling. Our work is defined within an uncertain framework. It is based on a conflict measure in the belief function theory applied between the messages of the different users of the thread. First of all, this method aims to analyze the messages relative to the posted question or topic. But detecting irrelevant message  is not enough to judge if this user is a troll or not. Thus, not only the content of the messages that can characterize the trolls but also their behaviors. 
Next, using the results of this analysis we measured the conflict between the different users. Finally, after calculating the conflict of each user we applied the kmeans method in order to distinguish trolls from the other users. Indeed, we have classified the users according to their conflict results into two clusters. This method was tested in different simulated data to check its feasibility. Since our proposed method for detecting malicious users dealt only with one discussion thread, we aim to extend this approach to detect trolls inside the community.



\ifCLASSOPTIONcaptionsoff
  \newpage
\fi



%

%


\vspace{-1cm}
\begin{IEEEbiographynophoto}{Imen Ouled Dlala}
 Univeristy of Tunis, LARODEC laboratory, High Institut of Management, Cite Bouchoucha, le Bardo, TUNISIA; E-mail:
dlalaimen@yahoo.fr
\end{IEEEbiographynophoto}
\vspace{-1cm}
\begin{IEEEbiographynophoto}{Dorra Attiaoui}
Univeristy of Tunis, LARODEC laboratory, High Institut of Management, Cite Bouchoucha, le Bardo, TUNISIA; University of Rennes 1, IRISA, Rue E. Branly, 22300 Lannion, FRANCE; E-mail: attiaoui.dorra@gmail.com
\end{IEEEbiographynophoto}
\vspace{-1cm}
\begin{IEEEbiographynophoto}{Arnaud Martin}
University of Rennes 1, IRISA, Rue E. Branly, 22300 Lannion, FRANCE; E-mail: Arnaud.Martin@univ-rennes1.fr
\end{IEEEbiographynophoto}
\vspace{-1cm}
\begin{IEEEbiographynophoto}{Boutheina Ben Yaghlane}
University of Carthage, IHEC Carthage, Carthage Presidence, TUNISIA, LARODEC laboratory; E-mail: boutheina.yaghlane@ihec.rnu.tn
\end{IEEEbiographynophoto}





\end{document}